\documentclass{article}

\usepackage[preprint]{neurips_2026}

\usepackage[utf8]{inputenc} 
\usepackage[T1]{fontenc}    
\usepackage{url}            
\usepackage{booktabs}       
\usepackage{amsfonts}       
\usepackage{amsmath}
\usepackage{nicefrac}       
\usepackage{microtype}      
\usepackage{xcolor}         
\usepackage{graphicx}
\usepackage{wrapfig}
\usepackage{algorithm}
\usepackage{enumitem}
\usepackage{algorithmic}
\usepackage{multicol}
\usepackage{multirow}
\usepackage{float} 
\usepackage{caption}
\usepackage[svgnames]{xcolor}
\usepackage[colorlinks=true, citecolor=DodgerBlue, linkcolor=Crimson, urlcolor=Crimson]{hyperref}

\newcommand{\etal}{\textit{et al}. }
\newcommand{\ie}{\textit{i}.\textit{e}., }
\newcommand{\eg}{\textit{e}.\textit{g}., }
\title{Disentangled Learning Improves Implicit Neural Representations for Medical Reconstruction}

\author{%
  Qing Wu\textsuperscript{1 2}\thanks{Most of this work was conducted during Ph.D. studies at ShanghaiTech University.}\quad Xuanyu Tian\textsuperscript{2}\quad  Chenhe Du\textsuperscript{2}\quad  \textbf{Haonan Zhang\textsuperscript{3}}\\\textbf{Xiao Wang\textsuperscript{2}}\quad \textbf{Le Lu\textsuperscript{1}}\quad \textbf{Yuyao Zhang\textsuperscript{2}}\thanks{Corresponding author.}\\
  \textsuperscript{1}Medical AI Lab, Ant Group\quad\textsuperscript{2}ShanghaiTech University\quad\textsuperscript{3}Shanghai Jiao Tong University \\
  \texttt{wu.qing@antgroup.com}\quad\texttt{zhangyy8@shanghaitech.edu.cn} \\
}

\begin{document}

\maketitle

\begin{abstract}
    Implicit neural representations (INRs) have emerged as a powerful paradigm for medical imaging via physics-informed unsupervised learning. Classical INRs optimize an entire network from scratch for each subject, leading to inefficient training and suboptimal imaging quality. Recent initialization-based approaches attempt to inject population priors into pre-trained networks, yet they rely on high-quality images and often suffer from catastrophic forgetting during fine-tuning. We present DisINR, a novel INR framework that explicitly disentangles shared and subject-specific representations. DisINR introduces a shared encoder–decoder pair and subject-specific encoders, whose features are jointly decoded for image reconstruction. By integrating differentiable forward models, it pre-trains the shared modules directly from limited raw measurements, removing the need for pre-acquired high-quality images. During test-time adaptation, only the subject-specific encoder is optimized, while the shared pair remains frozen, effectively preserving learned priors. Extensive evaluations on three representative medical imaging tasks show that DisINR significantly outperforms state-of-the-art INRs in both reconstruction accuracy and efficiency.
\end{abstract}

\section{Introduction}
\label{sec:intro}
\par Medical imaging is a cornerstone of modern clinical practice, enabling the detailed visualization of internal anatomical structures of the human body~\cite{harisinghani2019advances, rubin2014computed}. Inverse problems in medical imaging aim to recover anatomical images from raw measurements (\eg projections in CT or \textit{k}-space data in MRI). However, due to factors such as undersampling, these problems are inherently ill-posed with multiple suboptimal solutions~\cite{huang2024data, wang2023review}, calling for more advanced reconstruction algorithms.
\par Recently, implicit neural representations (INRs) have shown great potential in solving medical inverse problems through physics-informed unsupervised learning~\cite{molaei2023implicit, luo2025continuous}. INRs model each image as a continuous function parameterized by a coordinate-based neural network. With differentiable forward models (\eg Fourier transform for MRI), the network can be optimized without using external training data. Owing to the intrinsic bias of neural networks toward structured image patterns~\cite{rahaman2019spectral, mildenhall2021nerf}, INRs can resolve high-quality images in an unsupervised way. Yet, existing INR approaches~\cite{wu2023self,zha2022naf,feng2023imjense,cai2024structure,wu2025moner} typically optimize an entire network from scratch for each subject, which leads to low efficiency and limited reconstruction quality.
\par A promising direction is to learn an effective network initialization from pre-acquired high-quality images. Early studies~\cite{meta, lee2021meta, chen2022transformers} employ meta-learning, which performs well but requires a large number of pre-training images. More recently, Vyas~\etal\cite{STRAINER} proposed STRAINER, which learns transferable representations via a shared encoder and subject-specific decoders. This approach enables INRs to achieve a strong initialization even from just a few images. However, both types of initialization techniques face two key challenges: \textbf{\textit{1) Dependence on high-quality images.}} Such diagnosis-quality images are often difficult to obtain in medical scenarios, particularly for rare diseases; \textbf{\textit{2) Catastrophic forgetting.}} The population priors encoded in the initialized network are easily overwritten during case-specific fine-tuning, as all network parameters are updated at test time. These limitations reduce practical applicability and potential performance improvements. Moreover, existing initialization-based INR techniques are mainly developed for natural RGB images, which may limit their effectiveness on medical volumetric data.
\par In this work, we propose DisINR, a novel INR framework that explicitly disentangles shared and subject-specific representations. Specifically, DisINR introduces a shared encoder–decoder pair along with subject-specific encoders, whose features are jointly decoded to reconstruct images. By integrating differentiable forward models, DisINR can pre-train the encoder-decoder pair shared across diverse subjects directly from a limited set of raw measurements. This can effectively alleviate the reliance on high-quality images, significantly improving the model's applicability in a wide range of clinical scenarios. During test-time adaptation, we freeze the pre-trained pair, while optimizing only a subject-specific encoder for a given unseen test sample. This design fundamentally eliminates the catastrophic forgetting problem inherent in existing initialization techniques~\cite{meta,STRAINER}. Therefore, our method can effectively incorporate learned population priors into INR optimization, enabling faster and higher-quality reconstructions. We evaluate DisINR on three representative medical imaging tasks, including 3D volume fitting, undersampled MRI, and sparse-view CT. Extensive experiments show DisINR substantially outperforms state-of-the-art INRs in both accuracy and efficiency. 
\section{Background}
\paragraph{Medical Inverse Problems}
\par The data acquisition process of a medical imaging system can generally be formulated as below: $\mathbf{y}=\boldsymbol{A}\mathbf{x}+\boldsymbol{\epsilon}$, where $\mathbf{y}\in\mathbb{R}^m$ denotes measured data, $\mathbf{x}\in\mathbb{R}^n$ is the desired image, $\boldsymbol{A}\in\mathbb{R}^{m\times n}$ is the system matrix (\ie forward model), and $\boldsymbol{\epsilon}\in\mathbb{R}^m$ represents system noise. Note that the formulations are presented in the real domain for simplicity, but they also apply to complex-valued data, such as $\textit{k}$-space measurements in MRI. 
\par Medical inverse problems aim to reconstruct unknown images $\mathbf{x}$ from measurements $\mathbf{y}$. Due to various factors like undersampling (\ie $m \ll n$), these inverse problems are often highly ill-posed, where multiple suboptimal solutions exist. Conventional model-based algorithms~\cite{beister2012iterative,fessler2010model, thibault2007three} formulate it as the following optimization problem:
\begin{equation}
    \mathbf{x}^* = \underset{\mathbf{x}}{\arg\min}\ \|\boldsymbol{A}\mathbf{x}-\mathbf{y}\|_1+\lambda\cdot\mathcal{R}(\mathbf{x}),
\end{equation}
where $\mathbf{x}^*$ denotes the optimal solution, $\mathcal{R}$ is an explicit regularizer (\eg total variation~\cite{rudin1992nonlinearTV} for image smoothness), and $\lambda$ is a hyperparameter controlling the contribution of the regularizer. The use of the regularizer can effectively constrain the solution space, thus enabling improved image reconstructions. However, such handcrafted regularizers often fail to capture the complex distribution of medical images, thereby limiting the resulting image quality.
\begin{figure}[t]
    \centering
    \includegraphics[width=0.975\linewidth]{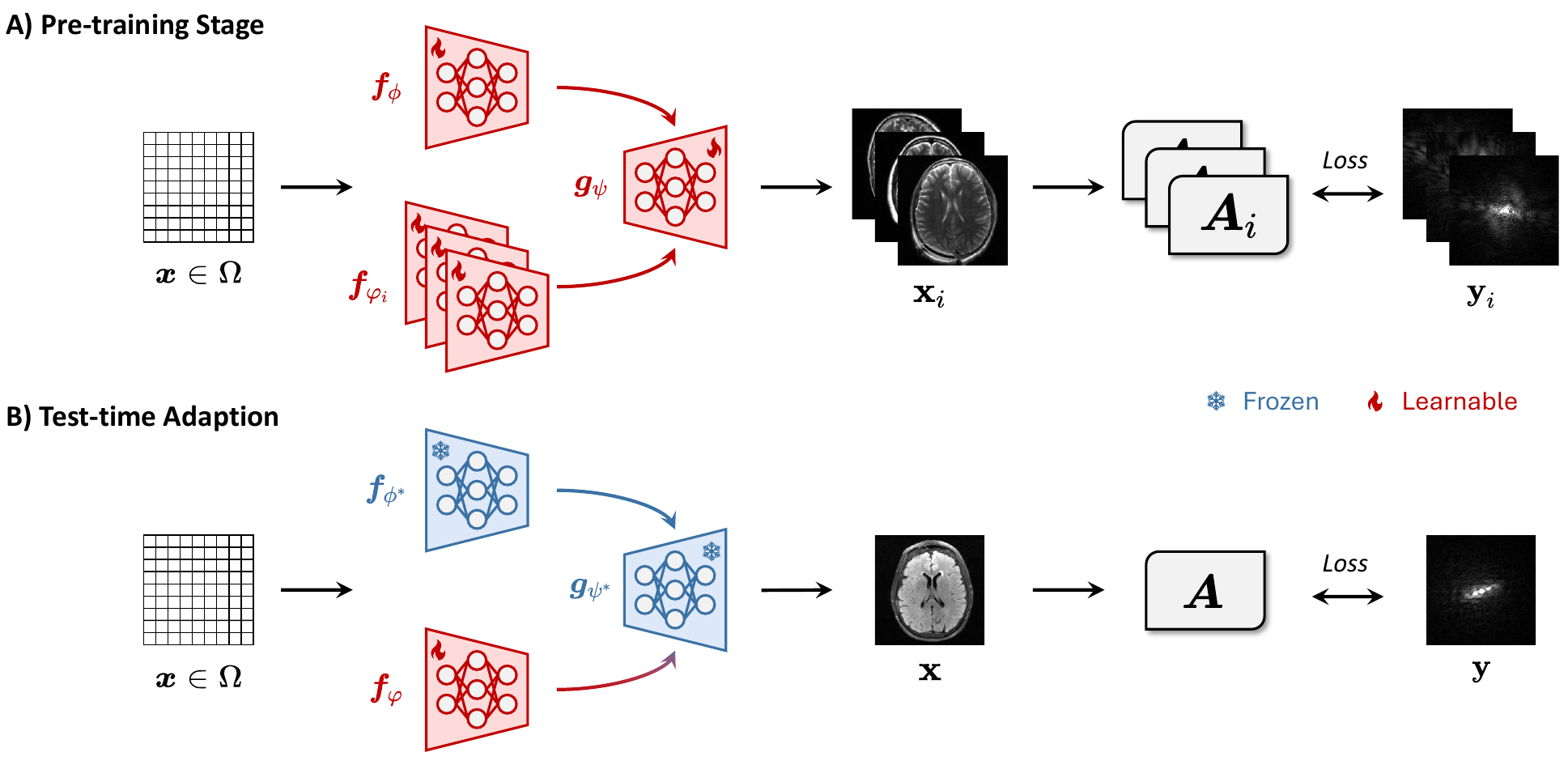}
    \caption{Overview of the proposed DisINR, which consists of a shared encoder–decoder pair $(\boldsymbol{f}_\phi,\boldsymbol{g}_\psi)$ and multiple subject-specific encoders $\{\boldsymbol{f}_{\varphi_i}\}_{i=1}^N$. \textbf{{A) Pre-training Stage:}} All modules in DisINR are jointly optimized from a few raw measurements $\mathbf{Y}=\{\mathbf{y}_i\}_{i=1}^N$ by incorporating differentiable forward models $\{\boldsymbol{A}_i\}_{i=1}^N$. \textbf{{B) Test-time Adaption:}} Given a new measurement $\mathbf{y}\notin\mathbf{Y}$ and the corresponding forward model $\boldsymbol{A}$, the shared pair $(\boldsymbol{f}_{\phi^*},\boldsymbol{g}_{\psi^*})$ is frozen while only the subject-specific encoder $\boldsymbol{f}_{\varphi}$ is optimized. This effectively injects learned population priors into the INR optimization process, enabling faster and higher-quality reconstruction $\mathbf{x}$.}
    \label{fig:fig_method_disINR}
\end{figure}
\paragraph{INR for Medical Reconstruction}
\label{sec:inr_for_medical_image}
\par As a signal representation way based on neural networks, implicit neural representations (INRs) have shown great potential in solving medical reconstruction, such as undersampled MRI~\cite{shen2022nerp, huang2023neural, feng2023imjense, liu2025physics, feng2025spatiotemporal, wu2025moner,xu2025accelerated, li2025universal} and CT~\cite{sun2021coil, zang2021intratomo, shen2022nerp, zha2022naf, wu2023self, wu2023unsupervised, cai2024structure, du2024dper}. Technically, INR models an image as a continuous function of coordinates: $\mathbf{c}\in\Omega \rightarrow \mathbf{x}$, and uses a coordinate-based neural network $\boldsymbol{h}_\theta$ to learn this mapping. Formally, INR solves the optimization problem:
\begin{equation}
    \theta^* = \underset{\theta}{\arg\min}\ \|\boldsymbol{A}\mathbf{x}-\mathbf{y}\|_1,\quad\mathrm{s.t.}\quad\mathbf{x}=\boldsymbol{h}_\theta(\mathbf{c}),
    \label{eq:inr}
\end{equation}
where $\theta^*$ denotes the optimized network parameters. Under this formulation, recovering the image $\mathbf{x}$ is transformed into optimizing the parameters $\theta$. Therefore, the desired image can be resolved as $\mathbf{x}^* = \boldsymbol{h}_{\theta^*}(\mathbf{c})$ in an unsupervised manner. The key insight is that neural networks possess an inherent and generalizable inductive bias toward continuous image structures~\cite{rahaman2019spectral,mildenhall2021nerf}, which serves as an implicit prior enabling high-quality reconstructions.
\paragraph{Advances for INR Initialization}
\par As shown in Eq.~\eqref{eq:inr}, classical INRs for medical reconstruction optimize an independent network from random Gaussian initialization, \ie $\theta\sim\mathcal{N}(0,1)$, for each subject. To enhance reconstruction efficiency and accuracy, recent works explore INR initialization strategies that generally fall into two categories: 1) Meta-learning~\cite{sitzmann2020metasdf, meta, lee2021meta, dupont2022data, chen2022transformers, kim2023generalizable,friedrich2025medfuncta}. For example, Tancik~\etal\cite{meta} showed that MAML- and Reptile-based meta-learning can significantly accelerate INR optimization and boost reconstruction quality across diverse tasks; 2) Transerable feature learning~\cite{STRAINER, vyas2025fit, rangarajan2025siedd, stolt2023nisf}. Vyas~\etal\cite{STRAINER} adopts a shared encoder and individual decoders, enabling feature transfer from a few high-quality images. However, these initialization methods depend on diagnosis-quality images and tend to overfit (\ie catastrophic forgetting) during test-time adaptation, severely limiting their applicability and performance.
\section{Proposed Method}
\paragraph{Overview}
\par Given a set $\mathbf{Y} = \{\mathbf{y}_i\}_{i=1}^N$ consisting of a limited number of undersampled measurements (\eg projections in CT or $\textit{k}$-space data in MRI), our goal is to reconstruct the corresponding high-quality image $\mathbf{x}$ from a new measurement $\mathbf{y} \notin \mathbf{Y}$ in a fully unsupervised way. This inverse problem involves two key challenges: {1)} How to extract rich population priors directly from the set $\mathbf{Y}$? {2)} How to inject the learned population priors into the model optimization?
\par To achieve this, we propose DisINR, a new INR framework that explicitly disentangles shared and subject-specific representations. Unlike conventional INR methods~\cite{feng2023imjense,cai2024structure} for medical reconstructions, which optimize an independent network $\boldsymbol{h}_\theta$ from scratch for each subject, as shown in Eq~\eqref{eq:inr}, our DisINR inherits a pre-trained encoder–decoder pair $(\boldsymbol{f}_{\phi^*}, \boldsymbol{g}_{\psi^*})$, while learning only a subject-specific encoder $\boldsymbol{f}_\varphi$ for the target. Formally, given a target $\mathbf{x}$, DisINR represents it as follows:
\begin{equation}
    \mathbf{x} = \boldsymbol{g}_{\psi^*}(\boldsymbol{f}_{\phi^*}(\mathbf{c})\odot \boldsymbol{f}_\varphi(\mathbf{c})),
\end{equation}
where $\odot$ denotes the concatenation operator, $\phi^*$ and $\psi^*$ represent the parameters of the pre-trained pair shared across diverse subjects, and $\varphi$ are the parameters of the encoder specialized for the target $\mathbf{x}$. By explicitly disentangling shared and subject-specific representations, our framework enables flexible transfer of the shared representation provided by the pre-trained encoder–decoder pair into subject-specific optimization during test-time adaptation. This design thus enables faster and improved reconstructions.
\begin{figure}[t] 
  \centering
  \begin{minipage}{0.48\textwidth}
    \input{algo1}
  \end{minipage}
  \hfill 
  \begin{minipage}{0.48\textwidth}
    \input{algo2}
  \end{minipage}
\end{figure}
\paragraph{Pre-training Stage}
\par Suppose the measurement set $\mathbf{Y}$ and the corresponding forward model set $\boldsymbol{A}=\{\boldsymbol{A}_i\}_{i=1}^N$, we first seek to embed the population priors inherent in the set $\mathbf{Y}$ into the shared encoder-decoder pair of DisINR. Fig.~\ref{fig:fig_method_disINR}\textbf{A} illustrates this pre-training stage. For each sample $\mathbf{y}_i \in \mathbf{Y}$, we first assign a specific encoder $\boldsymbol{f}_{\varphi_i}$.
Both the shared encoder $\boldsymbol{f}_\phi$ and the subject-specific encoder $\boldsymbol{f}_{\varphi_i}$ take the spatial coordinates at imaging space $\mathbf{c} \in \Omega$ as inputs, generating shared and subject-specific representations $\boldsymbol{f}_\phi(\mathbf{c})$ and $\boldsymbol{f}_{\varphi_i}(\mathbf{c})$, respectively. The decoder $\boldsymbol{g}_{\psi}$ then maps the concatenated features to the reconstructed image as $\mathbf{x}_i=\boldsymbol{g}_{\psi}(\boldsymbol{f}_{\phi}(\mathbf{c})\odot \boldsymbol{f}_{{\varphi}_i}(\mathbf{c}))$. Next, the predicted image $\mathbf{x}_i$ is projected to the measurement domain through the differentiable forward operator $\boldsymbol{A}_i$. Finally, we jointly optimize the three networks by minimizing the prediction error in the measurement domain. Mathematically, this process can be expressed as follows:
\begin{equation}
            \phi^*, \psi^*, \{\varphi^*_i\}_{i=1}^N = \underset{\phi, \psi, \{\varphi_i\}_{i=1}^N}{\arg\min}\ \sum_{i=1}^N\|\boldsymbol{A}_i\mathbf{x}_i-\mathbf{y}_i\|_1,\quad\mathrm{s.t.}\quad\mathbf{x}_i=\boldsymbol{g}_\psi(\boldsymbol{f}_\phi(\mathbf{c})\odot \boldsymbol{f}_{\varphi_i}(\mathbf{c})),
\end{equation}
where $\hat{\phi}$, $\hat{\psi}$, and $\{\hat{\varphi}_i\}_{i=1}^N$ denote learned parameters of the shared pair and the subject-specific encoders. The detailed procedure of the pre-training stage is presented in Algorithm~\ref{alg:pretrain}.
\paragraph{Test-time Adaptation}
\par After the model pre-training, the shared encoder $\boldsymbol{f}_{\phi^*}$ and decoder $\boldsymbol{g}_{\psi^*}$ capture rich representations that generalize across diverse subjects. During test-time adaptation, our goal is to recover the corresponding high-quality image $\mathbf{x}^*$ from a new measurement $\mathbf{y} \notin \mathbf{Y}$, without using any external data. As shown in Fig.~\ref{fig:fig_method_disINR}B, we freeze the shared encoder-decoder pair $(\boldsymbol{f}_{\phi^*},\boldsymbol{g}_{\psi^*})$, and optimize only a new subject-specific encoder $\boldsymbol{f}_\varphi$ from scratch, \ie $\varphi \sim \mathcal{N}(0, 1)$, by incorprating the differentiable forwad model $\boldsymbol{A}$. Formally, we solve the following optimization problem:
\begin{equation}
    \varphi^* = \underset{\varphi}{\arg\min}\ \|\boldsymbol{A\mathbf{x}}-\mathbf{y}\|_1,\quad\mathrm{s.t.}\quad\mathbf{x}=\boldsymbol{g}_{\psi^*}(\boldsymbol{f}_{\phi^*}(\mathbf{c})\odot \boldsymbol{f}_{\varphi}(\mathbf{c})),
\end{equation}
where $\varphi^*$ represents the learned parameters of the subject-specific encoder. Hence, the reconstruction is given by $\mathbf{x}^*=\boldsymbol{g}_{\psi^*}(\boldsymbol{f}_{\phi^*}(\mathbf{c})\odot \boldsymbol{f}_{\varphi^*}(\mathbf{c}))$. The procedure of the test-time adaptation is presented in Algorithm~\ref{alg:adapt}. By freezing the shared encoder-decoder pair, we can fundamentally mitigate the problem of catastrophic forgetting commonly observed in initialization-based techniques~\cite{meta,STRAINER}. This strategy effectively incorporates population priors into INR optimization, thereby enhancing reconstruction quality and efficiency.
\paragraph{Network Architecture}
\par Our DisINR framework is \textbf{\textit{architecture-agnostic}}, allowing it to be seamlessly integrated with different INR backbones. Owing to its efficient multi-resolution hash encoding, neural graphics primitive (NGP)~\cite{muller2022instant} has become one of the leading INR architectures for various medical imaging tasks, such as sparse-view CT~\cite{zha2022naf,cai2024structure,wu2023self} and undersampled MRI~\cite{feng2023imjense,feng2025spatiotemporal}. 
\par In this study, we thus use NGP as the backbone of DisINR. Both the shared encoder $\boldsymbol{f}_\phi$ and the subject-specific encoder $\boldsymbol{f}_{\varphi_i}$ share the same design, consisting of a multi-resolution hash encoding module followed by a two-layer MLP. The shared decoder $\boldsymbol{g}_\psi$ is also implemented as a two-layer MLP, which takes as input the concatenated features from the two encoders. Besides, \textbf{\textit{all network configurations are kept consistent across experiments}}, demonstrating the generalization and robustness of DisINR. Additional implementation details are provided in the Appendix.
\begin{figure}[t]
    \centering
    \includegraphics[width=\linewidth]{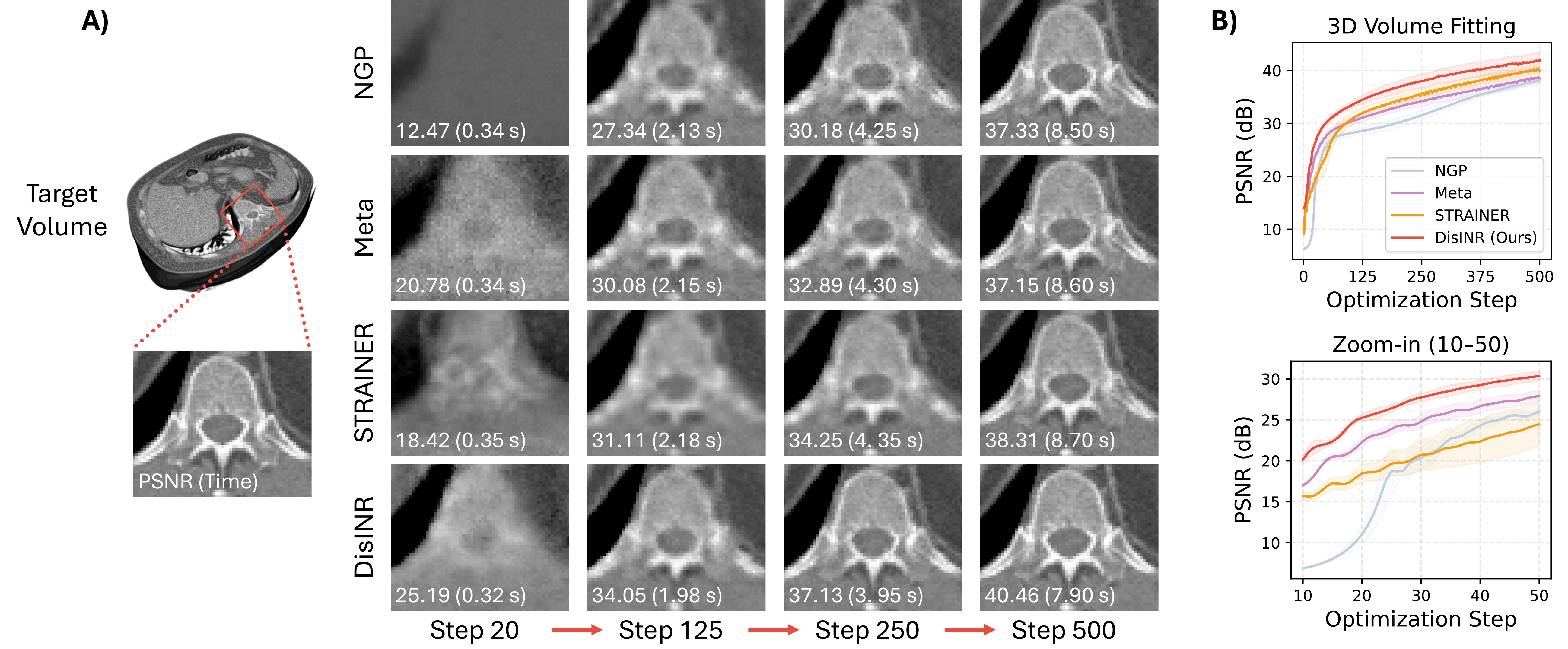}
    \caption{Qualitative result (\textbf{A}) on a representative case  and optimization curves (\textbf{B}) over all cases of three SOTA INR baselines and DisINR for 3D CT volume fitting from the AAPM dataset~\cite{aapm}.}
    \label{fig:fig_vf_img}
\end{figure}
\section{Experiments}
\par In this section, we evaluate the effectiveness and generalization of DisINR on three representative medical reconstruction tasks, including \textit{3D Volume Fitting}, \textit{Undersampled MRI}, and \textit{Sparse-view CT}. We also conduct extensive ablation studies to analyze the impact of key components in DisINR. \textbf{\textit{More details and additional results are provided in Appendix~\ref{sec:sec_app}.}}
\subsection{Experimental Settings}
\label{sec:exp_setting}
\paragraph{Dataset}
\par Our experiments involve multiple datasets across three tasks: 
\begin{itemize}[leftmargin=*]
    \item \underline{\textit{3D Volume Fitting.}} We employ 9 human body CT volumes from the AAPM dataset~\cite{aapm}, each with a size of 256$\times$256$\times$100. The volumes are first thresholded using a Hounsfield Unit (HU) window of [-800, 400] and then normalized to the range [0, 1]. Finally, we split these volumes into two subsets: 6 volumes for pre-training and 3 volumes for test. 
    \item \underline{\textit{Undersampled MRI.}} The fastMRI dataset~\cite{knoll2020fastmri} is one of the most commonly used public datasets for MRI reconstruction, containing large-scale multi-contrast \textit{k}-space data. Here, we use 150 multi-coil 2D brain \textit{k}-space samples of size 256$\times$256 from the fastMRI dataset~\cite{knoll2020fastmri}, including 100 T2w and 50 FLAIR scans. We simulate a 1D uniform Cartesian pattern with acceleration factors (AF) of 6 and 8. The dataset is divided into three subsets: 50 T2w samples for pre-training, 50 T2w samples for in-domain test, and 50 FLAIR samples for out-of-domain test.
    \item \underline{\textit{Spare-view CT.}} The DeepLesion~\cite{deeplesion} and LIDC~\cite{lidc} are two widely used public CT datasets, containing diverse human body scans collected from multiple clinical institutions. Specifically, we extract 100 and 50 slices of size 256$\times$256 from the two datasets, respectively. Each slice is thresholded using a HU window of [-800, 400] and normalized to the range [0, 1]. To generate raw projections, we simulate a 2D fan-beam geometry with 60 views using \texttt{CIL} toolbox~\cite{cil1,cil2}. The dataset is split into three parts: 50 cases from DeepLesion for pre-training, 50 cases from DeepLesion for in-domain test, and 50 cases from LIDC for out-of-domain test.
\end{itemize}
\paragraph{Baselines \& Metrics}
\par We compare DisINR with several representative baselines across three tasks, including both general-purpose INR methods and task-specific reconstruction approaches. \textbf{\textit{The general-purpose INR baselines}} include NGP~\cite{muller2022instant}, a grid-based INR architecture; Meta~\cite{meta}, a meta-learning-based initialization framework; and STRAINER~\cite{STRAINER}, an INR model for learning transferable representations. \textbf{\textit{For undersampled MRI}}, we include Zero-Filling (ZF), a classical reconstruction method, and IMJENSE~\cite{feng2023imjense}, a state-of-the-art INR-based model for parallel MRI. Since IMJENSE is originally built upon SIREN~\cite{siren}, we reproduce it using the NGP backbone~\cite{muller2022instant} for consistency. \textbf{\textit{For sparse-view CT}}, we include FBP, a standard analytical reconstruction algorithm, and SAX-NeRF~\cite{cai2024structure}, a recent INR-based method based on a line Transformer. We use the official implementation of SAX-NeRF in our experiments. For quantitative evaluation, we report peak signal-to-noise ratio (PSNR) and structural similarity index (SSIM) on all three tasks.
\subsection{Main Results}
\paragraph{3D Volume Fitting}
\begin{wraptable}{r}{0.45\textwidth}
    \centering

    \resizebox{0.45\textwidth}{!}{
    \setlength{\tabcolsep}{1mm}
    \begin{tabular}{lcccc}
    \toprule
    \multirow{2.5}{*}{\textbf{Method}} & \multicolumn{2}{c}{\textbf{Pre-training}} & \multicolumn{2}{c}{\textbf{Test}} \\ 
    \cmidrule(r){2-3}\cmidrule(r){4-5}
    & $\#$Param. & Time & $\#$Param. & Time \\ 
    \midrule
    NGP & \texttt{-} & \texttt{-} & 23.40 M & 8.5 s\\
    Meta & 23.40 M & 558.3 s & 23.40 M & 8.6 s\\
    STRAINER & 23.63 M & \textbf{124.9} s & 23.63 M & 8.7 s\\
    DisINR & \textbf{21.71 M} & 186.5 s & \textbf{10.83 M} & \textbf{7.9} s\\
    \bottomrule
    \end{tabular}}
    \caption{Comparison of learnable parameters and running time between three SOTA INR baselines and DisINR for 3D volume fitting on the AAPM dataset~\cite{aapm}.}
    \label{tab:tab_volume_fitting_training}
\end{wraptable}
\par Fig.~\ref{fig:fig_vf_img}\textbf{A} provides qualitative results. DisINR reconstructs clear anatomical structures with sharp tissue boundaries as early as step~125, whereas other methods still produce blurry results. Fig.~\ref{fig:fig_vf_img}\textbf{B} shows the optimization curves. DisINR converges significantly faster, reaching over 30~dB PSNR within 50 steps, while other methods remain around 25~dB. After full convergence (500 steps), it outperforms the second-best method STRAINER~\cite{STRAINER} by more than 2~dB, indicating that the learned shared representations provide a strong prior for both fast convergence and accurate reconstruction. 
\par In terms of efficiency, as summarized in Table~\ref{tab:tab_volume_fitting_training}, our DisINR maintains a compact model size (21.71~M parameters during pre-training) and reduces the number of learnable parameters to 10.83~M during test-time adaptation, since only the subject-specific encoder is optimized. This design also leads to the fastest inference speed (7.9~s). Instead, these baselines require updating the entire network at test time, resulting in higher computational cost and slower adaptation.
\begin{figure}[t]
    \centering
    \includegraphics[width=\linewidth]{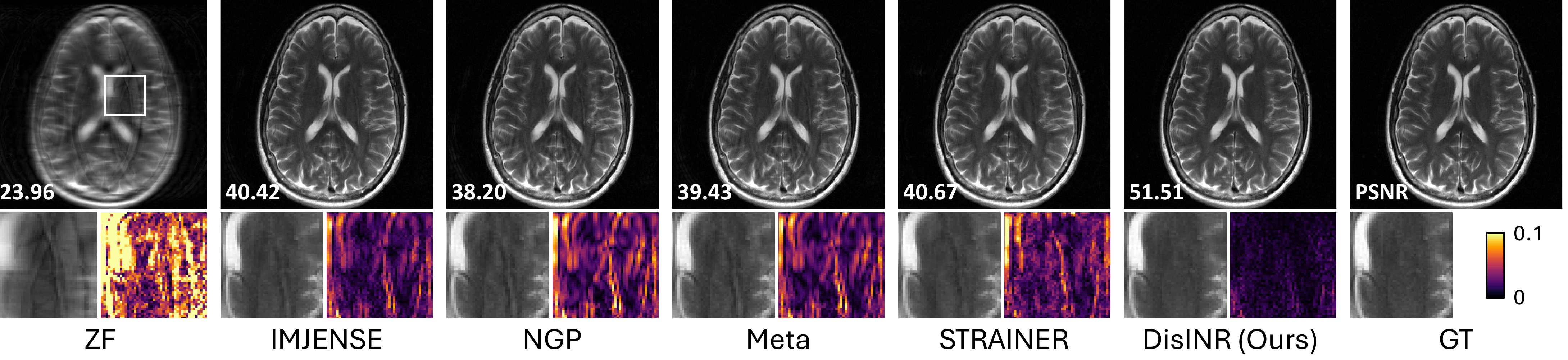}
    \caption{Quantitative comparison of five baselines and DisINR for undersampled MRI with a Cartesian pattern of AF = 6$\times$ on a representative sample of the fastMRI-T2w dataset~\cite{knoll2020fastmri}.}
    \label{fig:fig_result_mri}
\end{figure}
\begin{table}[t]
    \centering
    \resizebox{\textwidth}{!}{
    \setlength{\tabcolsep}{1mm}
    \begin{tabular}{lcccccccc}
    \toprule
    \multirow{4}{*}{\textbf{Method}} 
    & \multicolumn{4}{c}{\textbf{fastMRI-T2w (in-domain)}} 
    & \multicolumn{4}{c}{\textbf{fastMRI-FLAIR (out-of-domain)}} \\ 
    \cmidrule(r){2-5} \cmidrule(r){6-9}
     & \multicolumn{2}{c}{AF = 6$\times$}
     & \multicolumn{2}{c}{AF = 8$\times$}
     & \multicolumn{2}{c}{AF = 6$\times$}
     & \multicolumn{2}{c}{AF = 8$\times$} \\ 
    \cmidrule(r){2-3} \cmidrule(r){4-5} \cmidrule(r){6-7} \cmidrule(r){8-9}
     & PSNR & SSIM & PSNR & SSIM & PSNR & SSIM & PSNR & SSIM \\ 
    \midrule
     ZF          & 23.47{\scriptsize{$\pm$1.51}} & 0.681{\scriptsize{$\pm$0.036}} & 23.12{\scriptsize{$\pm$1.43}} & 0.660{\scriptsize{$\pm$0.036}} 
                           & 21.24{\scriptsize{$\pm$3.57}} & 0.684{\scriptsize{$\pm$0.045}} & 21.06{\scriptsize{$\pm$3.54}} & 0.655{\scriptsize{$\pm$0.048}} \\
     IMJENSE  & 38.76{\scriptsize{$\pm$5.28}} & 0.961{\scriptsize{$\pm$0.041}} & 29.30{\scriptsize{$\pm$2.84}} & 0.848{\scriptsize{$\pm$0.049}}
                           & 41.61{\scriptsize{$\pm$6.72}} & 0.979{\scriptsize{$\pm$0.022}} & 29.29{\scriptsize{$\pm$5.04}} & 0.901{\scriptsize{$\pm$0.055}} \\
     NGP    & 36.95{\scriptsize{$\pm$4.74}} & 0.947{\scriptsize{$\pm$0.043}} & 28.38{\scriptsize{$\pm$2.49}} & 0.818{\scriptsize{$\pm$0.047}}
                           & 41.36{\scriptsize{$\pm$7.59}} & 0.977{\scriptsize{$\pm$0.024}} & 28.87{\scriptsize{$\pm$5.10}} & 0.891{\scriptsize{$\pm$0.059}} \\
     Meta             & 37.42{\scriptsize{$\pm$4.78}} & 0.951{\scriptsize{$\pm$0.042}} & 28.56{\scriptsize{$\pm$2.57}} & 0.824{\scriptsize{$\pm$0.049}}
                           & 42.20{\scriptsize{$\pm$8.06}} & 0.980{\scriptsize{$\pm$0.023}} & 29.48{\scriptsize{$\pm$5.46}} & 0.900{\scriptsize{$\pm$0.059}} \\
     STRAINER        & 40.22{\scriptsize{$\pm$3.78}} & 0.972{\scriptsize{$\pm$0.032}} & 31.36{\scriptsize{$\pm$2.99}} & 0.892{\scriptsize{$\pm$0.041}}
                           & 37.28{\scriptsize{$\pm$5.17}} & 0.969{\scriptsize{$\pm$0.015}} & 28.17{\scriptsize{$\pm$4.90}} & 0.886{\scriptsize{$\pm$0.046}} \\
     DisINR                   & \textbf{48.42}{\scriptsize{$\pm$5.18}} & \textbf{0.992}{\scriptsize{$\pm$0.021}} & \textbf{35.46}{\scriptsize{$\pm$3.96}} & \textbf{0.952}{\scriptsize{$\pm$0.036}}
                           & \textbf{45.64}{\scriptsize{$\pm$4.79}} & \textbf{0.990}{\scriptsize{$\pm$0.009}} & \textbf{33.55}{\scriptsize{$\pm$4.82}} & \textbf{0.946}{\scriptsize{$\pm$0.026}} \\
    \bottomrule
    \end{tabular}}
    \caption{Quantitative results of five baselines and DisINR for undersampled MRI on the fastMRI-T2w and fastMRI-FLAIR datasets~\cite{knoll2020fastmri}.}
    \label{tab:tab_mri_double}
\end{table}

\paragraph{Undersampled MRI}
\par Table~\ref{tab:tab_mri_double} reports the quantitative comparisons. Overall, DisINR achieves the best performance across all AFs and domains. On the in-domain fastMRI-T2w dataset, DisINR produces 48.42~dB at AF = 6$\times$, outperforming STRAINER~\cite{STRAINER} by over 8~dB. On the out-of-domain fastMRI-FLAIR dataset, despite the contrast shift from the pre-training T2w data, DisINR still generalizes well, achieving 45.64~dB at AF = 6$\times$ and surpassing competing methods by 3--5~dB. These results demonstrate improved accuracy and generalization to unseen domains. Fig.~\ref{fig:fig_result_mri} shows qualitative results, where DisINR effectively removes undersampling artifacts while preserving both global structures and fine anatomical details, producing reconstructions closest to the GTs.
\par \textbf{\textit{We further evaluate robustness under different sampling trajectories.}} Models are pre-trained on fastMRI-T2w with Cartesian sampling and directly tested on radial (AF = 10$\times$) and Poisson (AF = 20$\times$) sampling without additional training. As shown in Table~\ref{tab:tab_sampling} and Fig.~\ref{fig:fig_sampling} of Appendix, DisINR consistently achieves the best performance, outperforming STRAINER by +3.34~dB and +2.07~dB in PSNR, respectively. This study further demonstrates the superiority of DisINR.
\begin{figure}[t]
    \centering
    \includegraphics[width=\linewidth]{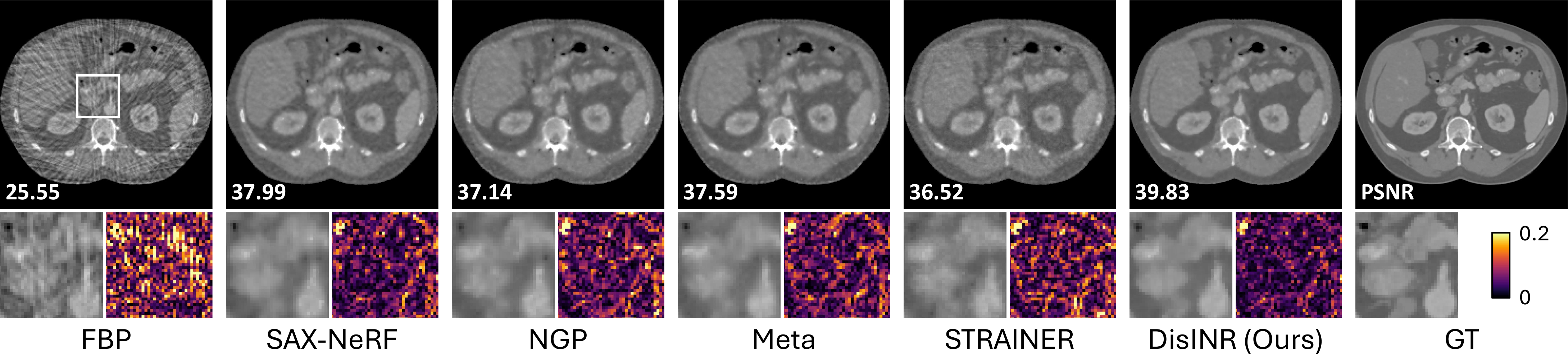}
    \caption{Qualitative results of five baselines and DisINR for sparse-view CT with 60 projection views on a representative sample of the DeepLesion dataset~\cite{deeplesion}.}
    \label{fig:fig_result_ct}
\end{figure}
\begin{table}[t]
    \centering
    \resizebox{\textwidth}{!}{
    \setlength{\tabcolsep}{1mm}
    \begin{tabular}{lcccccccc}
    \toprule
    \multirow{4}{*}{\textbf{Method}} 
    & \multicolumn{4}{c}{\textbf{DeepLesion (in-domain)}} 
    & \multicolumn{4}{c}{\textbf{LIDC (out-of-domain)}} \\ 
    \cmidrule(r){2-5} \cmidrule(r){6-9}
     & \multicolumn{2}{c}{60 Views} 
     & \multicolumn{2}{c}{90 Views} 
     & \multicolumn{2}{c}{60 Views} 
     & \multicolumn{2}{c}{90 Views}  \\ 
    \cmidrule(r){2-3} \cmidrule(r){4-5} \cmidrule(r){6-7} \cmidrule(r){8-9}
     & PSNR & SSIM & PSNR & SSIM & PSNR & SSIM & PSNR & SSIM \\ 
    \midrule
     FBP          & 23.98{\scriptsize{$\pm$1.25}} & 0.362{\scriptsize{$\pm$0.039}} 
                           & 26.70{\scriptsize{$\pm$1.25}} & 0.478{\scriptsize{$\pm$0.043}} 
                           & 24.47{\scriptsize{$\pm$2.03}} & 0.403{\scriptsize{$\pm$0.062}} & 27.40{\scriptsize{$\pm$2.39}} & 0.524{\scriptsize{$\pm$0.076}} \\
     SAX-NeRF  & 37.71{\scriptsize{$\pm$1.60}} & 0.965{\scriptsize{$\pm$0.010}} 
                           & 38.92{\scriptsize{$\pm$1.49}} & 0.975{\scriptsize{$\pm$0.007}}
                           & 37.34{\scriptsize{$\pm$3.00}} & 0.948{\scriptsize{$\pm$0.036}} & 38.13{\scriptsize{$\pm$3.02}} & 0.956{\scriptsize{$\pm$0.033}} \\
     NGP    & 36.89{\scriptsize{$\pm$1.42}} & 0.957{\scriptsize{$\pm$0.010}} 
                           & 38.09{\scriptsize{$\pm$1.34}} & 0.969{\scriptsize{$\pm$0.007}}
                           & 36.50{\scriptsize{$\pm$2.73}} & 0.940{\scriptsize{$\pm$0.036}} & 37.50{\scriptsize{$\pm$2.74}} & 0.951{\scriptsize{$\pm$0.033}} \\
     Meta             & 36.96{\scriptsize{$\pm$1.35}} & 0.958{\scriptsize{$\pm$0.010}} 
                           & 38.08{\scriptsize{$\pm$1.34}} & 0.969{\scriptsize{$\pm$0.007}}
                           & 36.61{\scriptsize{$\pm$2.72}} & 0.941{\scriptsize{$\pm$0.036}} & 37.52{\scriptsize{$\pm$2.69}} & 0.951{\scriptsize{$\pm$0.032}} \\
     STRAINER        & 36.17{\scriptsize{$\pm$1.46}} & 0.945{\scriptsize{$\pm$0.014}} 
                           & 37.05{\scriptsize{$\pm$1.40}} & 0.956{\scriptsize{$\pm$0.011}}
                           & 35.35{\scriptsize{$\pm$2.77}} & 0.920{\scriptsize{$\pm$0.043}} & 36.17{\scriptsize{$\pm$2.76}} & 0.932{\scriptsize{$\pm$0.039}} \\
     DisINR                   & \textbf{39.24}{\scriptsize{$\pm$1.65}} & \textbf{0.976}{\scriptsize{$\pm$0.008}} 
                           & \textbf{40.30}{\scriptsize{$\pm$1.54}} & \textbf{0.982}{\scriptsize{$\pm$0.006}}
                           & \textbf{38.02}{\scriptsize{$\pm$2.97}} & \textbf{0.954}{\scriptsize{$\pm$0.034}} & \textbf{38.85}{\scriptsize{$\pm$2.98}} & \textbf{0.962}{\scriptsize{$\pm$0.031}} \\
    \bottomrule
    \end{tabular}}
    \caption{Quantitative results of five baselines and DisINR for sparse-view CT on the DeepLesion~\cite{deeplesion} and LIDC~\cite{lidc} datasets.}
    \label{tab:tab_svct}
\end{table}

\paragraph{Sparse-View CT}
\par Table~\ref{tab:tab_svct} shows the quantitative results. We observe that DisINR achieves the highest PSNR and SSIM in all settings, outperforming the second-best baseline with 2--3 dB in PSNR. We also provide the qualitative comparison in Fig.~\ref{fig:fig_result_ct}. From the visualization, our DisINR recovers the sharper anatomical details while reducing streak artifacts. 
\par \textbf{\textit{We also use a public COVID-19 CT dataset~\cite{shakouri2021covid19} to evaluate model performance on abnormal anatomical structures.}} The dataset contains lung CT images from patients with COVID-19. We directly test five baselines and our DisINR, all pre-trained on the DeepLesion dataset~\cite{deeplesion}, on this unseen dataset. As shown in Table~\ref{tab:tab_covid19} of Appendix, DisINR achieves the best performance, significantly outperforming STRAINER, the second-best method, by +2 dB in PSNR. We provide qualitative results in Fig.~\ref{fig:fig_covid19} of Appendix, where DisINR produces clearer images compared to baselines. Overall, the evaluation confirms the robustness of our framework on the structural anomalies. 
\begin{figure}[h]
    \centering
    \includegraphics[width=\linewidth]{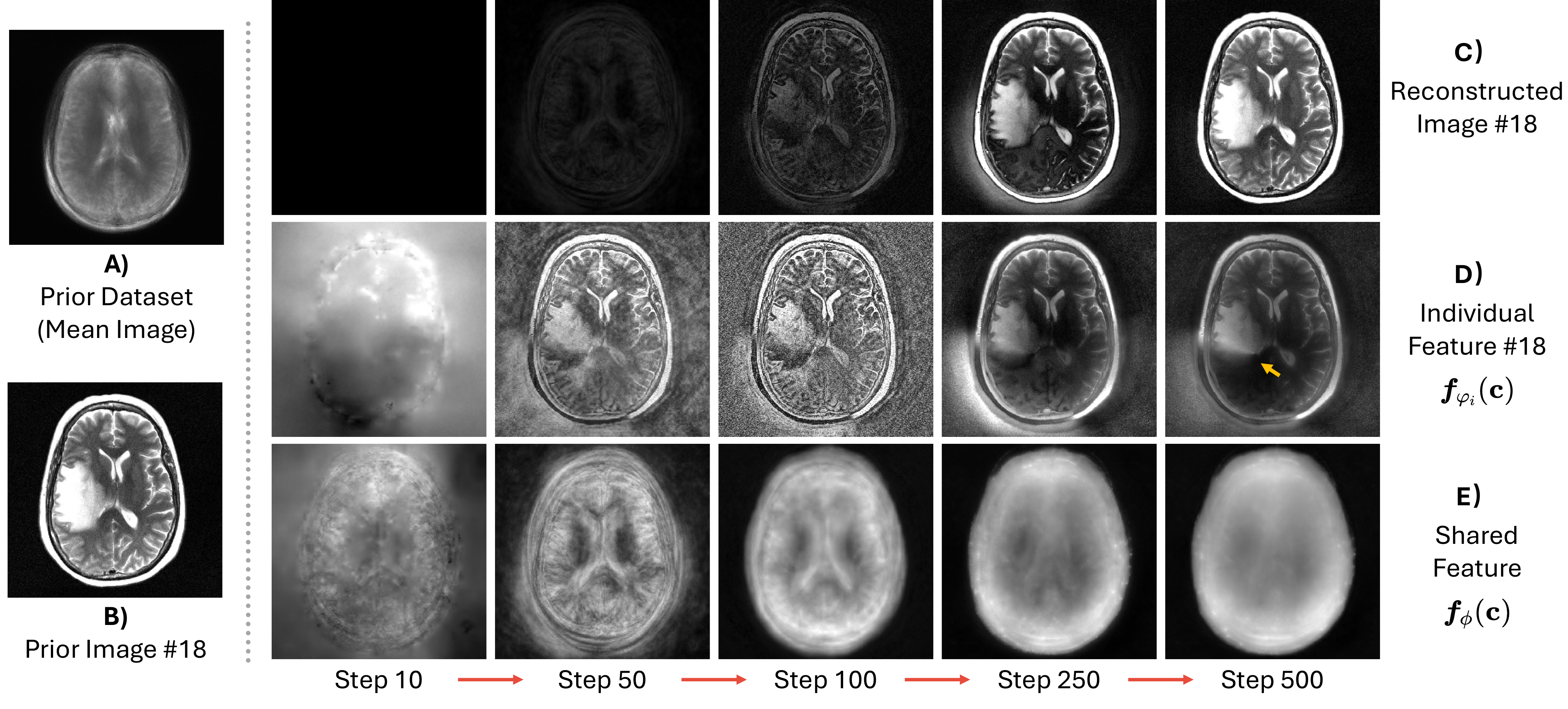}
    \caption{Visualization of DisINR pretraining on the fastMRI-T2w dataset~\cite{knoll2020fastmri}: \textbf{A)} Prior dataset, mean image of all prior samples. \textbf{B)} Prior image $\#$18. \textbf{C)} Reconstructed image $\#$18 by DisINR. \textbf{D)} Individual feature extracted by the subject-specific encoder of DisINR for sample $\#$18. \textbf{E)} Shared feature extracted by the shared encoder of DisINR from the prior dataset.}
    \label{fig:fig_disten}
\end{figure}
\subsection{Discussion}
\paragraph{Visualization and Analysis of Disentangled Learning}
\par The strong performance gains of DisINR are primarily driven by its ability to learn disentangled representations. To better understand this property, we conduct experiments on the fastMRI-T2w dataset~\cite{knoll2020fastmri} and examine the shared and subject-specific representations during pre-training. Specifically, we apply PCA to the features $\boldsymbol{f}_\phi(\mathbf{c})$ and $\boldsymbol{f}_{\varphi_i}(\mathbf{c})$, and visualize the resulting principal components.
\par The visualizations are shown in Fig.~\ref{fig:fig_disten}. We make the following key observations: 1) the shared representation $\boldsymbol{f}_\phi(\mathbf{c})$ captures smooth, population-level information; 2) the shared representation is not simply the mean image, but instead encodes higher-level semantic structures; and 3) the individual features $\boldsymbol{f}_{\varphi_i}$ capture subject-specific details, such as \textit{\textbf{the hemorrhage regions}}, as indicated by the yellow arrow. This study demonstrates that DisINR can effectively extract and disentangle population-level representations, offering empirical evidence of its robustness and strong generalization capability.
\paragraph{Effect of Weight freezing in DisINR}
\par
\begin{figure}[h]
    \centering
    \includegraphics[width=0.9\textwidth]{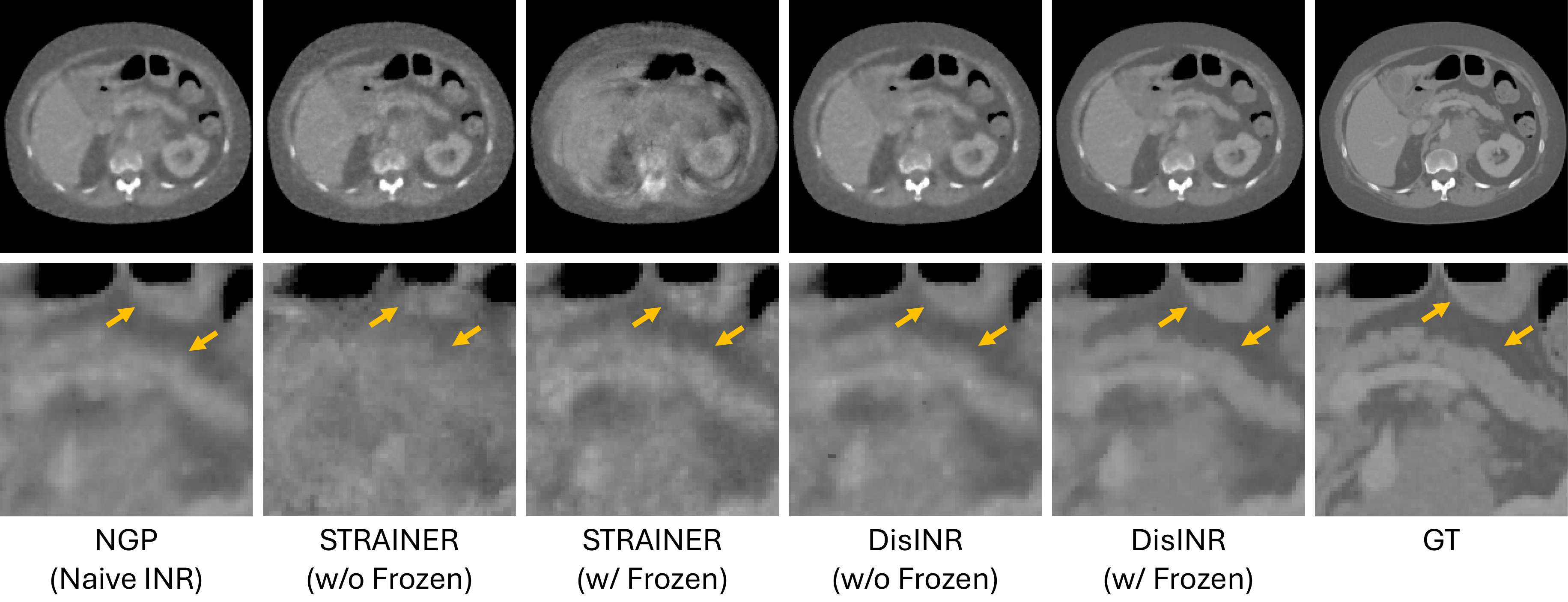}
    \caption{Qualitative results of two baselines and DisINR ablating frozen components for sparse-view CT with 60 projection views on a representative sample of the DeepLesion dataset~\cite{deeplesion}.}
    \label{fig:fig_frozen}
\end{figure}

 To study the effect of weight freezing on model performance, we perform an ablation study using the DeepLesion dataset~\cite{deeplesion}. As shown in Table~\ref{tab:tab_frozen}, we make three observations: 1) Freezing the shared components in DisINR slightly improves performance (+1.47 dB in PSNR). 2) Regardless of freezing, DisINR consistently outperforms existing SOTA INR-based methods (NGP~\cite{muller2022instant} and STRAINER~\cite{STRAINER}). 3) Freezing the shared components in STRAINER~\cite{STRAINER} significantly reduces performance (about -8 dB in PSNR). Fig.~\ref{fig:fig_frozen} shows the qualitative results, where DisINR (w/ Frozen) consistently recovers the best reconstructions.
\par We attribute these results to the following factors: 1) Without freezing the shared encoder-decoder, DisINR is prone to catastrophic forgetting, which degrades performance; 2) By learning disentangled representations with separate encoders, DisINR obtains more effective semantic representations, enabling better reconstruction even without freezing shared components; and 3) STRAINER’s single-encoder design limits its ability to separate shared and subject-specific representations, resulting in ineffective transfer and requiring full fine-tuning, consistent with its original formulation. Overall, this study validates the effectiveness of weight freezing in our DisINR.
\begin{table}
    \centering
    \resizebox{0.825\textwidth}{!}{
    \begin{tabular}{l l c c}
    \toprule
    \textbf{Method} & \textbf{Frozen Components} & \textbf{PSNR} & \textbf{SSIM} \\
    \midrule
    
    NGP (Naive INR)       & N/A                    & 37.30$\pm${\scriptsize 1.20} & 0.960$\pm${\scriptsize 0.009} \\
    STRAINER (w/o Frozen) & N/A                    & 36.65$\pm${\scriptsize 1.24} & 0.950$\pm${\scriptsize 0.012} \\
    STRAINER (w/ Frozen)  & Shared Encoder         & 28.40$\pm${\scriptsize 2.34} & 0.832$\pm${\scriptsize 0.057} \\
    DisINR (w/o Frozen)   & N/A                    & 38.18$\pm${\scriptsize 1.12} & 0.967$\pm${\scriptsize 0.007} \\
    DisINR (w/ Frozen)    & Shared Encoder-Decoder Pair & \textbf{39.63$\pm${\scriptsize 1.39}} & \textbf{0.977$\pm${\scriptsize 0.006}} \\
    
    \bottomrule
    \end{tabular}}
    \caption{Quantitative results of two baselines and DisINR ablating frozen components for sparse-view CT with 60 projection views on the DeepLesion dataset~\cite{deeplesion}.}
    \label{tab:tab_frozen}
\end{table}
\paragraph{Effect of Network Backbone}
\begin{table}[t]
\centering
\resizebox{0.8\textwidth}{!}{
\begin{tabular}{l c c c c c c}
\toprule
\multirow{2.5}{*}{\textbf{Method}} 
& \multicolumn{2}{c}{\textbf{NeRF}} 
& \multicolumn{2}{c}{\textbf{SIREN}} 
& \multicolumn{2}{c}{\textbf{NGP}} \\
\cmidrule(r){2-3} \cmidrule(r){4-5} \cmidrule(r){6-7}
& PSNR & $\#$Param. & PSNR & $\#$Param. & PSNR & $\#$Param. \\
\midrule

Naive INR 
& 37.03$\pm${\scriptsize 2.23} & 0.34M 
& 35.73$\pm${\scriptsize 1.15} & 0.46M 
& 38.73$\pm${\scriptsize 5.56} & 4.92M \\

STRAINER 
& 36.78$\pm${\scriptsize 1.61} & 0.34M 
& 35.76$\pm${\scriptsize 2.54} & 0.43M 
& 41.75$\pm${\scriptsize 2.33} & 4.97M \\

DisINR 
& \textbf{40.16$\pm${\scriptsize 1.96}} & \textbf{0.31M} 
& \textbf{38.26$\pm${\scriptsize 1.04}} & 0.44M 
& \textbf{49.61$\pm${\scriptsize 3.82}} & 4.94M \\

\bottomrule
\end{tabular}}
\caption{Quantitative results of vanilla INR, STRAINER~\cite{STRAINER}, and DisINR using different backbones for undersampled MRI with a Cartesian pattern of AF = 6$\times$ on the fastMRI-T2w dataset~\cite{knoll2020fastmri}.}
\label{tab:tab_network}
\end{table}


    
    
    
    
    
    
\par To evaluate the robustness of our DisINR across different backbones, we implement vanilla INR, STRAINER~\cite{STRAINER}, and DisINR using three representative architectures (NeRF~\cite{mildenhall2021nerf}, SIREN~\cite{siren}, and NGP~\cite{muller2022instant}). For a fair comparison, all methods are configured with comparable numbers of learnable parameters under each backbone, and we carefully tune hyperparameters for all baselines. Table~\ref{tab:tab_network} shows the quantitative results. First, DisINR consistently outperforms competing methods across all backbones, demonstrating robustness to architectural choices. Second, all methods benefit from the stronger NGP backbone, which provides higher capacity and improves reconstruction performance. Fig.~\ref{fig:fig_network} of Appendix shows qualitative results.
\paragraph{Effect of Pre-training Data Size}
\par To study how the size of the pre-training sample affects reconstruction quality, we conduct experiments on the fastMRI-T2w dataset~\cite{knoll2020fastmri} using a Cartesian pattern with AF = 6$\times$. We vary the number of available pre-training subjects ($N$ = 5, 10, 50) and compare our DisINR with IMJENSE~\cite{feng2023imjense} and STRAINER~\cite{STRAINER}. Note that IMJENSE~\cite{feng2023imjense} does not involve any pre-training stage and thus serves as a fixed baseline.
\par
\begin{wrapfigure}{r}{0.325\textwidth}
    \centering
    \includegraphics[width=\linewidth]{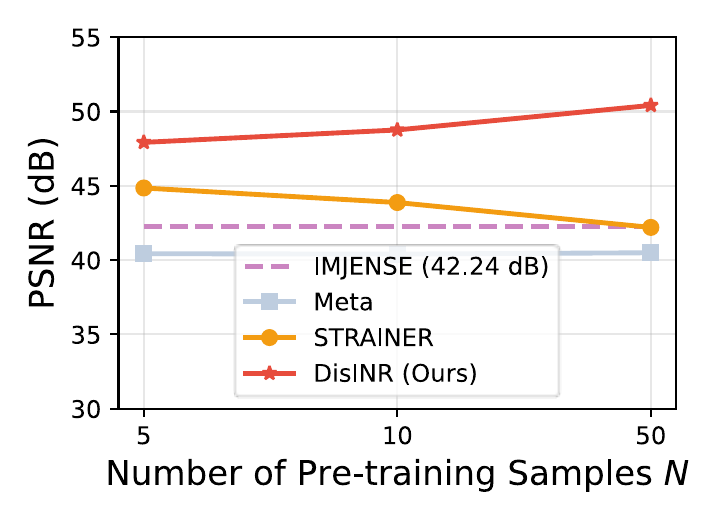}
    \caption{Performance curves of three SOTA INR baselines and DisINR under different numbers of pre-training samples $N$ for undersampled MRI with AF = 6$\times$ on the fastMRI-T2w dataset~\cite{knoll2020fastmri}.}
    \label{fig:fig_number_psnr}
\end{wrapfigure}

 Fig.~\ref{fig:fig_number_psnr} reports the quantitative results. our DisINR consistently achieves the highest PSNR under all settings, surpassing IMJENSE~\cite{feng2023imjense} and STRAINER~\cite{STRAINER} by a clear margin, and its performance increases monotonically with $N$ from 47.92 dB (5 samples) to 50.41 dB (50 samples). In contrast, STRAINER shows no gain from more pre-training data and even degrades slightly (44.85 to 42.20 dB), indicating the limited scalability of its single-shared encoder design. To sum up, this study confirms DisINR's scalability and its ability to effectively extract and reuse population priors for robust reconstruction.
\section{Conclusion \& Limitation}
\label{sec:conclusion}
\par In this work, we present DisINR, a new architecture-agnostic framework to incorporate population priors into INR learning for medical reconstruction. Using physics-informed unsupervised learning, DisINR effectively pre-trains a shared encoder–decoder pair to capture rich representations directly from a few raw measurements. By freezing the learned encoder–decoder pair, a subject-specific encoder is then optimized for each new subject. Evaluations on three classic medical imaging tasks show DisINR achieves SOTA performance compared to existing INR techniques in both accuracy and efficiency.
\par A key limitation of this work is the limited exploration of scalability with respect to the size of the pre-training dataset. Although we conduct a preliminary study on datasets ranging from 5 to 50 subjects, a more comprehensive evaluation on larger-scale data is still needed. Nevertheless, DisINR already achieves strong performance with relatively small pre-training sets ($N \leq$ 50), confirming the effectiveness of its disentangled representation in capturing population-level features. More extensive large-scale validation is left for future work.

\bibliography{ref}
\bibliographystyle{abbrv}

\clearpage
\appendix

\section{Appendix}
\label{sec:sec_app}
\subsection{Experimental Details}
\label{sec:app_exp_det}
\paragraph{Data Pre-processing}
\par In our experiments, we include two classical medical imaging tasks: undersampled MRI and sparse-view CT. Here, we detail the data pre-processing. \textbf{\textit{We will release our code upon acceptance to improve reproducibility.}}
\begin{figure}[h]
    \centering
    \includegraphics[width=0.925\linewidth]{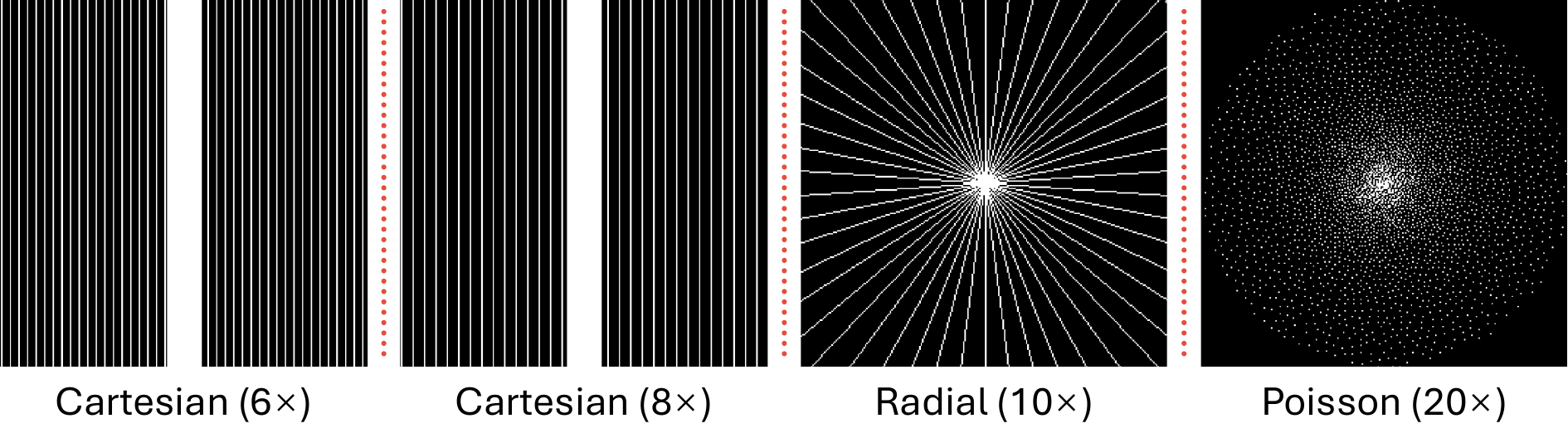}
    \caption{Four types of sampling patterns used in the undersampled MRI task.}
    \label{fig:fig_sm_mri_pattern}
\end{figure}
\begin{itemize}[leftmargin=*]
\item \underline{\textit{Undersampled MRI.}} Given multi-coil 2D brain \textit{k}-space data of size 256$\times$256 from the fastMRI dataset~\cite{knoll2020fastmri}, as shown in Fig.~\ref{fig:fig_sm_mri_pattern}, we simulate 1D Cartesian sampling patterns with acceleration factors of 6 and 8, where the size of the auto-calibration region (ACS) is set to 24. While coil-sensitivity maps computed from the raw fully sampled \textit{k}-space data are used for parallel MRI.
\item \underline{\textit{Sparse-view CT.}} For raw 2D slices from the DeepLesion~\cite{deeplesion} and LIDC~\cite{lidc} datasets, we generate sparse-view CT projections by using the CIL toolbox~\cite{cil1,cil2} to simulate a 2D fan-beam geometry. The detailed acquisition parameters are provided in Table~\ref{tab:geometry_2d}.
\end{itemize}
\begin{table}[h]
    \centering
    \resizebox{0.5\textwidth}{!}{
    \setlength{\tabcolsep}{1mm}
    \begin{tabular}{lc}
    \toprule
    \textbf{Parameters} & \textbf{Values} \\
    \midrule
    Type of geometry & 2D fan-beam \\
    Image Size & 256$\times$256 \\
    Voxel Size (mm\textsuperscript{2}) & 1$\times$1 \\
    View Range ($^\circ$) & [0, 360) \\
    Number of Detectors & 500 \\
    Detector Spacing (mm) & 2 \\
    Number of Projection Views & 60/90 \\
    Distance from Source to Center (mm) & 300 \\
    Distance from Center to Detector (mm) & 300 \\
    \bottomrule
    \end{tabular}}
    \vspace{2mm}
    \caption{Acquisition parameters of 2D fan-beam geometry used in the sparse-view CT task.}
    \label{tab:geometry_2d}
\end{table}

\paragraph{Quantitative Metrics}
\par In our evaluation, we use peak signal-to-noise ratio (PSNR) and structural similarity index (SSIM)~\cite{ssim}, two widely used visual metrics, to quantitatively assess the model performance. Specifically, for undersampled MRI reconstruction, we calculate these two metrics using the normalized amplitude map, as MRI images are complex-valued. For CT reconstruction (including 3D volume fitting and sparse-view CT), we directly compute them using the reconstructed CT images without applying additional pre-processing.
\paragraph{Implementations of DisINR}
\par The proposed DisINR framework is \textit{\textbf{architecture-agnostic}}, allowing it to be seamlessly integrated with different INR backbones. In this study, we adopt NGP~\cite{muller2022instant} as the backbone of DisINR. Specifically, both the shared encoder $f_\phi$ and the subject-specific encoder $f_{\varphi_i}$ are composed of a hash encoding module followed by two fully connected (FC) layers, each containing 128 neurons with ReLU activation. The output dimension of each encoder is 128. The shared decoder $g_\psi$ consists of two FC layers: the first layer has 128 hidden units with ReLU activation, and the output layer is linear (without activation). The decoder takes as input the concatenated feature vector from the shared and subject-specific encoders, resulting in an input dimension of 256. For the hash encoding, we use the following configuration: number of levels $L=10$, number of entries per level $T=2^{18}$, feature dimension $F=8$, minimum resolution $N_{\text{min}}=2$, and per-level factor $b=2$. 
\par For model optimization, we employ the Adam optimizer~\cite{kingma2014adam} with its default hyperparameters. The learning rate is initialized to $1\times10^{-3}$ and decayed by a factor of 0.5 every 1{,}000 iterations. The network is trained for a total of 4{,}000 iterations. \textbf{\textit{Note that the same optimization configurations are used for both the pre-training and test-time adaptation stages. All configurations are kept consistent across all experiments}}, further demonstrating the generalization and robustness of DisINR. 
\subsection{Results on Unseen MRI Sampling Patterns}
\begin{table}[h]
\centering
\resizebox{0.675\textwidth}{!}{
\begin{tabular}{l c c c c}
\toprule
\multirow{2.5}{*}{\textbf{Method}} 
& \multicolumn{2}{c}{\textbf{Radial (AF = 10$\times$)}} 
& \multicolumn{2}{c}{\textbf{Poisson (AF = 20$\times$)}} \\
\cmidrule(r){2-3} \cmidrule(r){4-5}
& PSNR & SSIM & PSNR & SSIM \\
\midrule

ZF          & 18.84$\pm${\scriptsize 1.95} & 0.520$\pm${\scriptsize 0.056} & 16.57$\pm${\scriptsize 1.53} & 0.402$\pm${\scriptsize 0.043} \\
IMJENSE     & 34.51$\pm${\scriptsize 3.42} & 0.968$\pm${\scriptsize 0.010} & 34.78$\pm${\scriptsize 3.92} & 0.951$\pm${\scriptsize 0.014} \\
NGP         & 31.94$\pm${\scriptsize 3.50} & 0.926$\pm${\scriptsize 0.024} & 30.99$\pm${\scriptsize 3.38} & 0.843$\pm${\scriptsize 0.055} \\
Meta     & 32.01$\pm${\scriptsize 3.52} & 0.924$\pm${\scriptsize 0.026} & 30.97$\pm${\scriptsize 3.58} & 0.841$\pm${\scriptsize 0.063} \\
STRAINER    & 33.88$\pm${\scriptsize 3.20} & 0.960$\pm${\scriptsize 0.011} & 34.88$\pm${\scriptsize 3.71} & 0.955$\pm${\scriptsize 0.015} \\
DisINR 
            & \textbf{37.22$\pm${\scriptsize 2.79}} & \textbf{0.976$\pm${\scriptsize 0.006}} 
            & \textbf{36.95$\pm${\scriptsize 3.88}} & \textbf{0.964$\pm${\scriptsize 0.011}} \\

\bottomrule
\end{tabular}}
\caption{Quantitative results of five baselines and DisINR for undersampled MRI with unseen radial (AF = 10$\times$) and Poisson patterns (AF = 20$\times$) on the fastMRI-T2w dataset~\cite{knoll2020fastmri}.}
\label{tab:tab_sampling}
\end{table}
\begin{figure}[h]
    \centering
    \includegraphics[width=\linewidth]{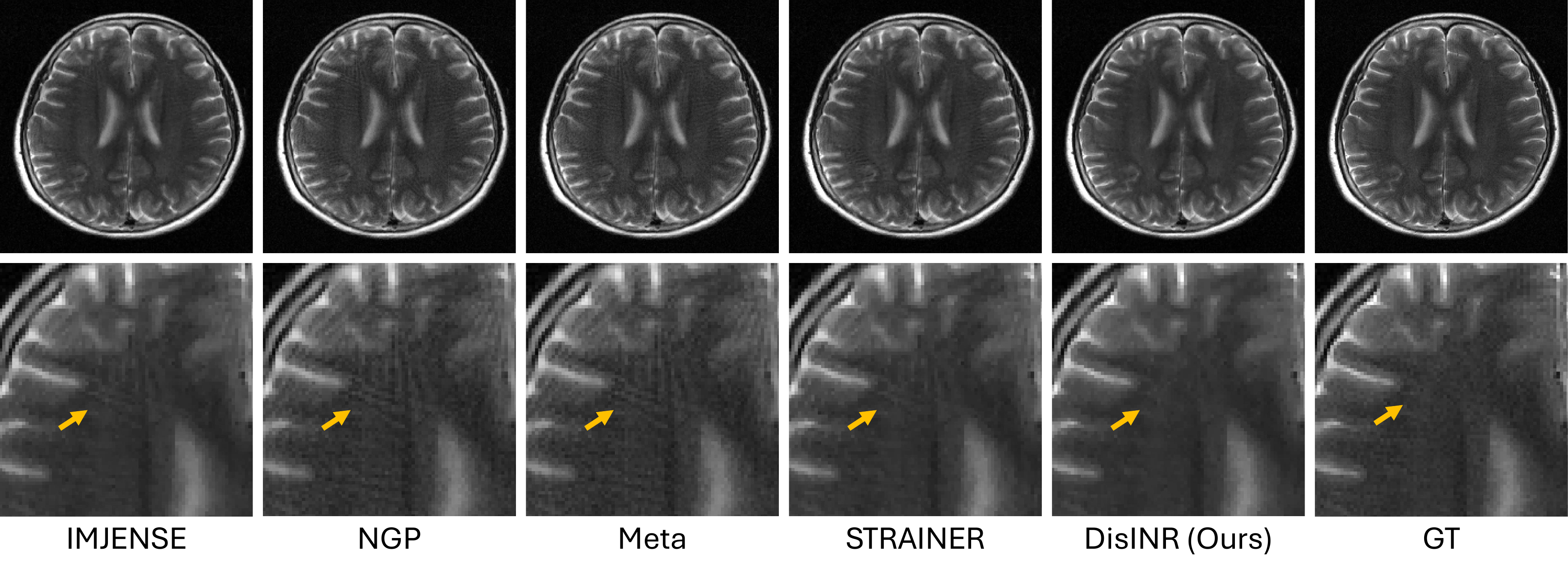}
    \caption{Qualitative results of four baselines and DisINR for undersampled MRI with a radial pattern of AF = 10$\times$ on a representative sample of the fastMRI-T2w dataset~\cite{knoll2020fastmri}.}
    \label{fig:fig_sampling}
\end{figure}
\subsection{Results on Unseen Abnormal Anatomical Structures}
\begin{table}[h]
\centering
\resizebox{0.7\linewidth}{!}{
\begin{tabular}{l c c c c}
\toprule
\multirow{2.5}{*}{\textbf{Method}} 
& \multicolumn{2}{c}{\textbf{60 Views}} 
& \multicolumn{2}{c}{\textbf{90 Views}} \\
\cmidrule(r){2-3} \cmidrule(r){4-5}
& PSNR & SSIM & PSNR & SSIM \\
\midrule

FBP         & 18.95$\pm${\scriptsize 0.90} & 0.280$\pm${\scriptsize 0.011} & 21.87$\pm${\scriptsize 0.90} & 0.358$\pm${\scriptsize 0.017} \\
SAX-NeRF   & 32.09$\pm${\scriptsize 1.39} & 0.926$\pm${\scriptsize 0.003} & 33.02$\pm${\scriptsize 1.36} & 0.940$\pm${\scriptsize 0.003} \\
NGP         & 31.50$\pm${\scriptsize 1.30} & 0.916$\pm${\scriptsize 0.004} & 32.45$\pm${\scriptsize 0.97} & 0.933$\pm${\scriptsize 0.005} \\
Meta     & 31.55$\pm${\scriptsize 1.23} & 0.916$\pm${\scriptsize 0.004} & 32.43$\pm${\scriptsize 1.03} & 0.932$\pm${\scriptsize 0.005} \\
STRAINER    & 30.68$\pm${\scriptsize 1.27} & 0.899$\pm${\scriptsize 0.006} & 31.45$\pm${\scriptsize 1.11} & 0.916$\pm${\scriptsize 0.003} \\
DisINR 
            & \textbf{32.73$\pm${\scriptsize 1.43}} & \textbf{0.935$\pm${\scriptsize 0.005}} 
            & \textbf{33.72$\pm${\scriptsize 1.18}} & \textbf{0.948$\pm${\scriptsize 0.005}} \\

\bottomrule
\end{tabular}}
\caption{Quantitative results of five baselines and DisINR for sparse-view CT with 60 and 90 projection views on the unseen COVID-19 dataset~\cite{shakouri2021covid19} containing abnormal anatomical structures.}
\label{tab:tab_covid19}
\end{table}
\begin{figure}[t]
    \centering
    \includegraphics[width=\linewidth]{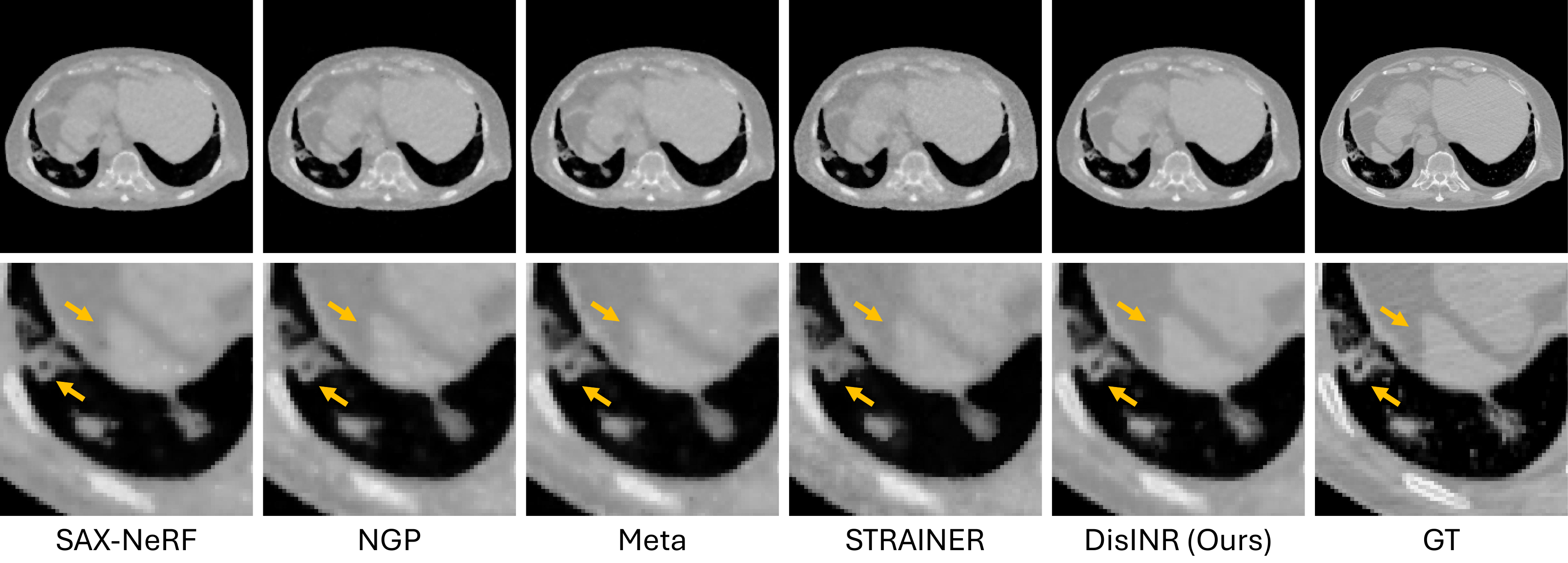}
    \caption{Qualitative results of four baselines and DisINR for sparse-view CT with 60 projection views on a representative sample of the unseen COVID-19 dataset~\cite{shakouri2021covid19}.}
    \label{fig:fig_covid19}
\end{figure}
\begin{figure}
    \centering
    \includegraphics[width=\linewidth]{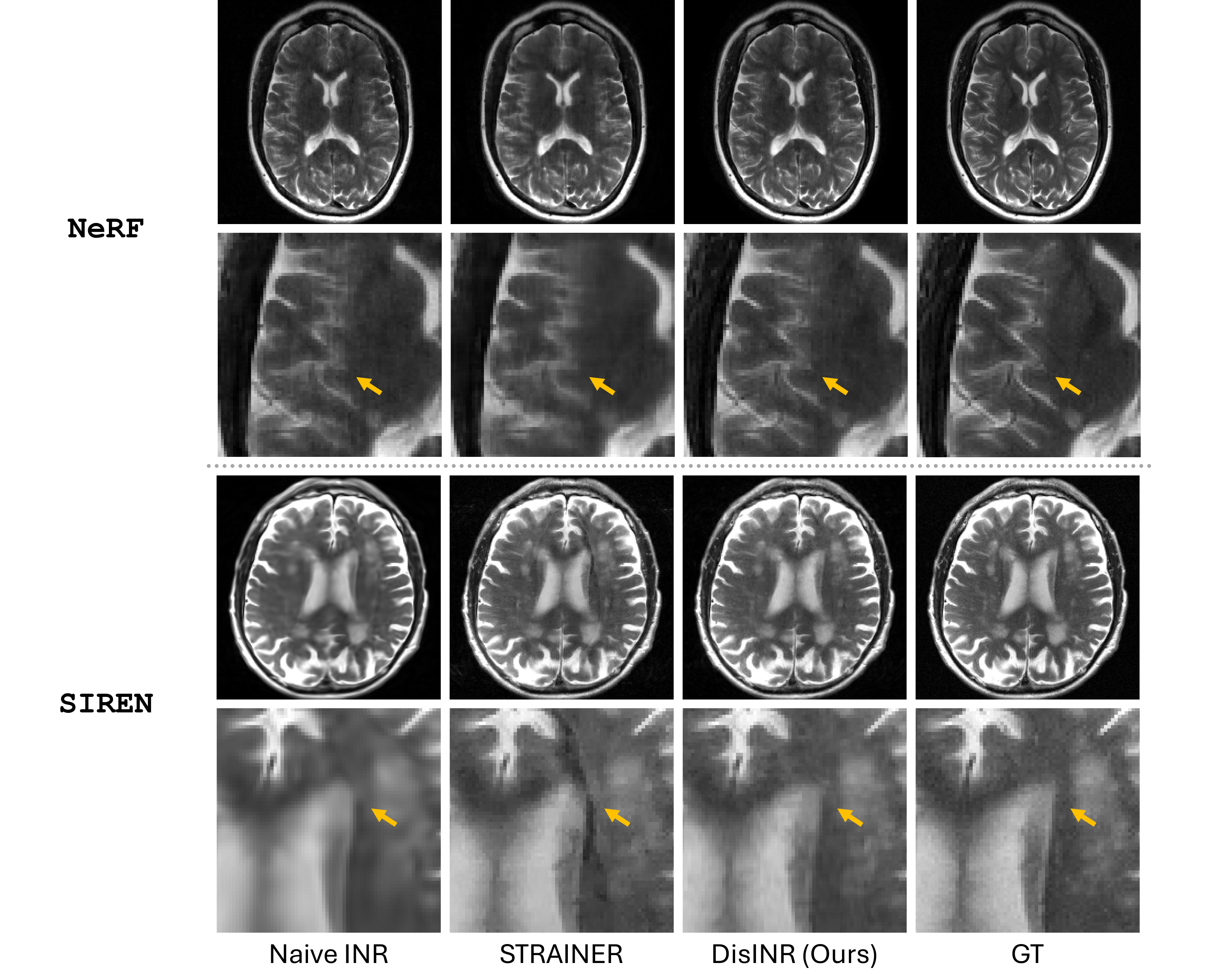}
    \caption{Qualitative results of two baselines and DisINR with different INR architectures (NeRF and SIREN) for undersampled MRI with a Cartesian pattern of AF = 6$\times$ on a representative sample of the fastMRI-T2w dataset~\cite{knoll2020fastmri}.}
    \label{fig:fig_network}
\end{figure}
\clearpage
\subsection{Additional Visual Results}
\par
\begin{figure*}[h]
    \centering
    \includegraphics[width=\linewidth]{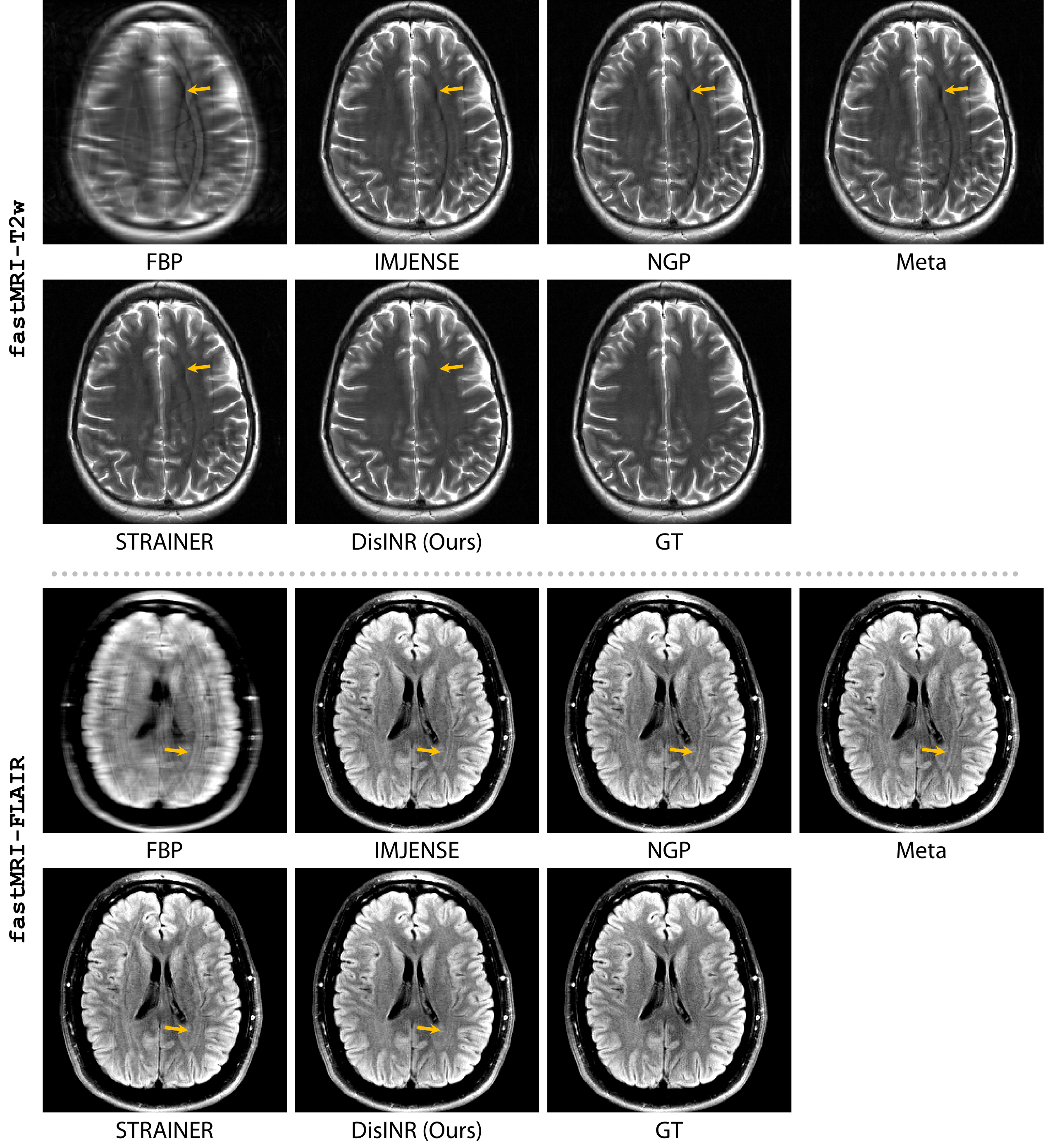}
    \caption{Quantitative comparison of five baselines and our DisINR for undersampled MRI with a Cartesian pattern of AF = 6$\times$ on two representative samples of the fastMRI-T2w and fastMRI-FLAIR datasets~\cite{knoll2020fastmri}.}
    \label{fig:fig_sm_mri}
\end{figure*}

\begin{figure*}[h]
    \centering
    \includegraphics[width=\linewidth]{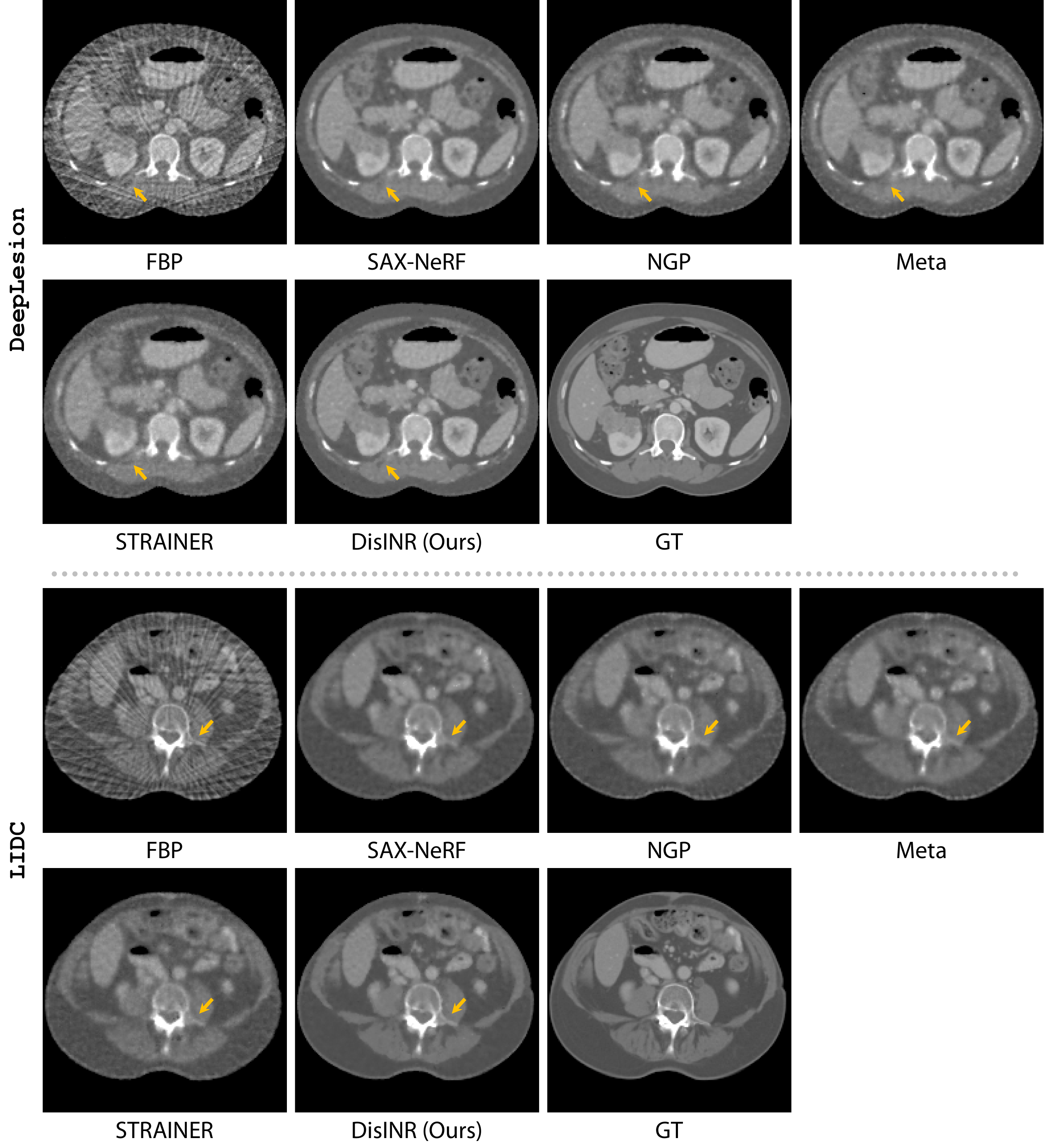}
    \caption{Qualitative results of five baselines and our DisINR for sparse-view CT with 60 projection views on two representative samples of the DeepLesion~\cite{deeplesion} and LIDC~\cite{lidc} datasets.}
    \label{fig:fig_sm_svct}
\end{figure*}

 Fig.~\ref{fig:fig_sm_svct} and Fig.~\ref{fig:fig_sm_mri} show additional Qualitative results between our DisINR and the baselines for the undersampled MRI and sparse-view CT tasks. Visually, our DisINR produces the best reconstructions, which are closest to the GT samples.

\end{document}